\documentclass{article}
\PassOptionsToPackage{numbers, sort&compress}{natbib}
\usepackage[preprint]{neurips_2023} 

\usepackage{framed, multirow, cuted}
\usepackage{booktabs}

\usepackage{amssymb}
\usepackage{subcaption}
\usepackage{latexsym}

\usepackage[utf8]{inputenc} %
\usepackage[T1]{fontenc}    %
\usepackage[backref=page]{hyperref}       %
\usepackage{url}            %
\usepackage{booktabs}       %
\usepackage{amsfonts}       %
\usepackage{nicefrac}       %
\usepackage{microtype}      %
\usepackage{xcolor}         %
\definecolor{mydarkblue}{rgb}{0,0.08,0.55}
\definecolor{DarkRed}{HTML}{780000}
\definecolor{RegRed}{HTML}{C1121F}
\definecolor{CoolBlue}{HTML}{669BBC}

\hypersetup{  %
    colorlinks=true,
    linkcolor=mydarkblue,
    citecolor=mydarkblue,
    filecolor=mydarkblue,
    urlcolor=mydarkblue
}

\usepackage{hyperref}
\usepackage{amsmath}
\usepackage{algorithm}
\usepackage{algpseudocode}
\usepackage{listings}
\usepackage{xspace}
\usepackage{wrapfig}
\usepackage{enumitem}

\usepackage[textsize=scriptsize, textwidth=32mm]{todonotes}
\newcommand{\authoredtodo}[4][]{\ifhmode\unskip\fi\todo[linecolor=#3, backgroundcolor=#3!50!white,#1]{[#2] #4}}
\definecolor{turquoise}{rgb}{0.0, 0.6, 0.6}

\definecolor{deepmagenta}{rgb}{0.8, 0.0, 0.8}

\definecolor{apricot}{rgb}{1, 0.8, 0.7}

\definecolor{burgundy}{rgb}{0.5, 0.0, 0.125}

\definecolor{newcolor}{rgb}{.8,.349,.1}

\newcommand{\CoxPH}{Cox\xspace}

\newcommand{\dataSet}{\mathcal{N}}
\newcommand{\uncensSet}{\mathcal{N}_{\text{uncens}}}
\newcommand{\censSet}{\mathcal{N}_{\text{cens}}}
\newcommand{\leftCensSet}{\mathcal{N}_{\text{left-cens}}}
\newcommand{\intervalCensSet}{\mathcal{N}_{\text{interval-cens}}}
\newcommand{\tmax}{T_{\text{max}}}
\newcommand{\method}{CenTime\xspace}

\newcommand{\eg}{\textit{e.g.}\xspace}
\newcommand{\ie}{\textit{i.e.}\xspace}
\newcommand{\beq}{\begin{equation}}
\newcommand{\eeq}{\end{equation}}

\newcommand{\cb}[1]{\left\{ {#1} \right\}}
\newcommand{\br}[1]{\left( {#1} \right)}
\newcommand{\sq}[1]{\left[ {#1} \right]}

\newcommand\numberthis{\addtocounter{equation}{1}\tag{\theequation}}

\newcommand{\figref}[1]{Fig. \ref{#1}}
\newcommand{\appref}[1]{Appendix \ref{#1}}
\newcommand{\secref}[1]{Sec. \ref{#1}}
\newcommand{\tabref}[1]{Table \ref{#1}}
\renewcommand{\eqref}[1]{Eq. \ref{#1}}

\usepackage[detect-all]{siunitx}
\usepackage{graphicx}
\usepackage{comment}

\title{\method: Event-Conditional Modelling of Censoring in Survival Analysis}%

\author{
    Ahmed H. Shahin$^{1,2}$ \quad An Zhao$^{1}$\quad Alexander C. Whitehead$^{2}$\quad Daniel C. Alexander$^{1}$ \and \textbf{Joseph Jacob$^{1,3}$}\quad \textbf{David Barber$^{2}$}\\
    $^{1}$Centre for Medical Image Computing, University College London, London, UK\\
    $^{2}$Centre for Artificial Intelligence, University College London, London, UK\\
    $^{3}$Lungs for Living Research Centre, University College London, London, UK\\
}
   
\begin{document}
\maketitle

\begin{abstract}
Survival analysis is a valuable tool for estimating the time until specific events, such as death or cancer recurrence, based on baseline observations. This is particularly useful in healthcare to prognostically predict clinically important events based on patient data. However, existing approaches often have limitations; some focus only on ranking patients by survivability, neglecting to estimate the actual event time, while others treat the problem as a classification task, ignoring the inherent time-ordered structure of the events.
Furthermore, the effective utilization of censored samples$-$training data points where the exact event time is unknown$-$is essential for improving the predictive accuracy of the model.
In this paper, we introduce \method, a novel approach to survival analysis that directly estimates the time to event.
Our method features an innovative event-conditional censoring mechanism that performs robustly even when uncensored data is scarce. We demonstrate that our approach forms a consistent estimator for the event model parameters, even in the absence of uncensored data. Furthermore, \method is easily integrated with deep learning models with no restrictions on batch size or the number of uncensored samples. We compare our approach with standard survival analysis methods, including the Cox proportional-hazard model and DeepHit. Our results indicate that \method offers state-of-the-art performance in predicting time-to-death while maintaining comparable ranking performance. Our implementation is publicly available at \url{https://github.com/ahmedhshahin/CenTime}.
\end{abstract}

\section{Introduction}
\label{intro}
Survival analysis has been applied in many areas, including genomics \cite{lee2019}, healthcare \cite{lee2018,zhao2022prognostic,shahin22,lu2023}, manufacturing \cite{richardeau2012}, marketing \cite{kim2014}, and social sciences \cite{emmert2019introduction}. To keep the language concrete, we will discuss and evaluate the healthcare scenario of patient survival, bearing in mind that our methods are generally applicable. 

Survival analysis has a long research history, from traditional statistical methods to modern machine learning methods \cite{wang2019machine}. Kaplan and Meier proposed an early method that models the proportions of patients at risk at given times \cite{kaplan1958nonparametric}. The main constraint of the K-M approach is that it cannot model the influence of covariates. The later developed Cox proportional hazards model \cite{cox72} overcomes some of the limitations of the K-M approach, but cannot directly estimate survival times. Rather, it estimates the relative likelihood of death for one patient compared to another.

A key challenge in survival analysis is dealing with censored data. In right censoring, we know that the patient was alive up to the censoring time, but we do not know when they died or indeed if they are still alive. Naively disregarding censored samples negatively impacts the performance of survival models and leads to statistically biased results \cite{BUCKLEY1979LinearData}. As a result, a large body of the literature has focused on leveraging censored training samples to improve survival models' performance and make more accurate predictions.

Our main contribution is to directly model the time of death of a patient, for which we introduce a novel censoring model. We compare our approach to the standard Cox model \cite{cox72,katzman2018}, DeepHit \cite{lee2018}, and the classical censoring model and apply these methods to predicting the survival of Idiopathic Pulmonary Fibrosis (IPF) patients based on volumetric Computerized Tomography (CT) images and associated clinical data. IPF is a progressive fibrotic lung disease with a variable and unpredictable progression rate, making it an ideal test case for our proposed approaches. For completeness, we include comparisons with the Cox model, using standard techniques to estimate actual survival time from a ranking.

\subsection{Preliminaries}
We aim to learn the distribution $p_\theta(D=t|x)$, where $D$ is a random variable associated with the death time $t$; $\theta$ represents model parameters and $x$ is a set of covariates (\eg CT scan and clinical data). For simplicity, we assume that the death time $t\in\cb{1,\ldots,\tmax}$ is discrete and refers to the number of months a patient survives post the CT scan.

Our training data $\mathcal{D}$ is a collection of uncensored and right-censored observations. The observation for an uncensored sample is represented as $(\delta_n=1, x_n, t_n)$, where $\delta_n=1$ indicates that the death time $t_n$ is known. For a right-censored sample, the observation is represented as $(\delta_n=0, x_n, c_n)$, where $\delta_n=0$ indicates that the death time $t_n$ is unknown, and only the censoring time $c_n<t_n$ is known, $n=1,\ldots, N$. The data index set is $\dataSet$, the uncensored observation index set is $\uncensSet = \{n:\delta_n=1\}$, and the censored observation index set is $\censSet = \{n:\delta_n=0\}$. 
\section{\method: an event-conditional censoring model}
We introduce \method which enables the direct learning of a death time distribution $p_\theta(D=t|x)$ from either censored or uncensored data. \method uses a novel censoring mechanism that we believe is more representative of censoring in some clinical situations. The method is generally applicable to other forms of censoring (left, interval), see \appref{app:additional_censoring}. Here we concentrate on right censoring. Specifically, we first sample the death time and then generate a censoring time from a distribution up to the death time. This results in the censored time model
\beq
p_\theta(C=c|x) = \sum_{t=1}^{\tmax} p(C=c|D=t,x)p_\theta(D=t|x)
\label{eq:our_censoring_general}
\eeq
The objective then is to maximize the log likelihood
\begin{align*}
    \mathbb{L}(\theta)
    &\equiv \sum_{n\in \uncensSet} \log{p_\theta(D=t_n|x_n)} + \sum_{i\in \censSet} {\log{p_\theta(C=c_i|x_i)}} \numberthis
    \label{eq:new:one}
\end{align*}
Objective in \eqref{eq:new:one} is the likelihood of a mixture model containing contributions from the uncensored data and censored data, with each term being a consistent objective for the event model parameters $\theta$ (\ie estimators based on either contribution converge to the true parameters as the number of samples increases). This implies that even in the scenario where we only have censored training data, the model can learn the underlying event model.

The model also has the advantage that, if needed, we can easily sample data from this model a given proportion of censored to uncensored data. If a proportion of censored to uncensored data $p_c:p_n$ is required, for a chosen $N$ one can simply sample $Np_c$ censored datapoints from $p_\theta(C=c_n|x_n)$ and $Np_n$ uncensored datapoints from $p_\theta(D=t_n|x_n)$. This feature is absent in classical censoring models, in which it is not possible to sample data with a required proportion of censored to uncensored data. 
\begin{figure*}[t]
    \hspace{0.025\textwidth}
    \begin{subfigure}{0.45\textwidth}
        \centering 
        \includegraphics[width=0.9\textwidth]{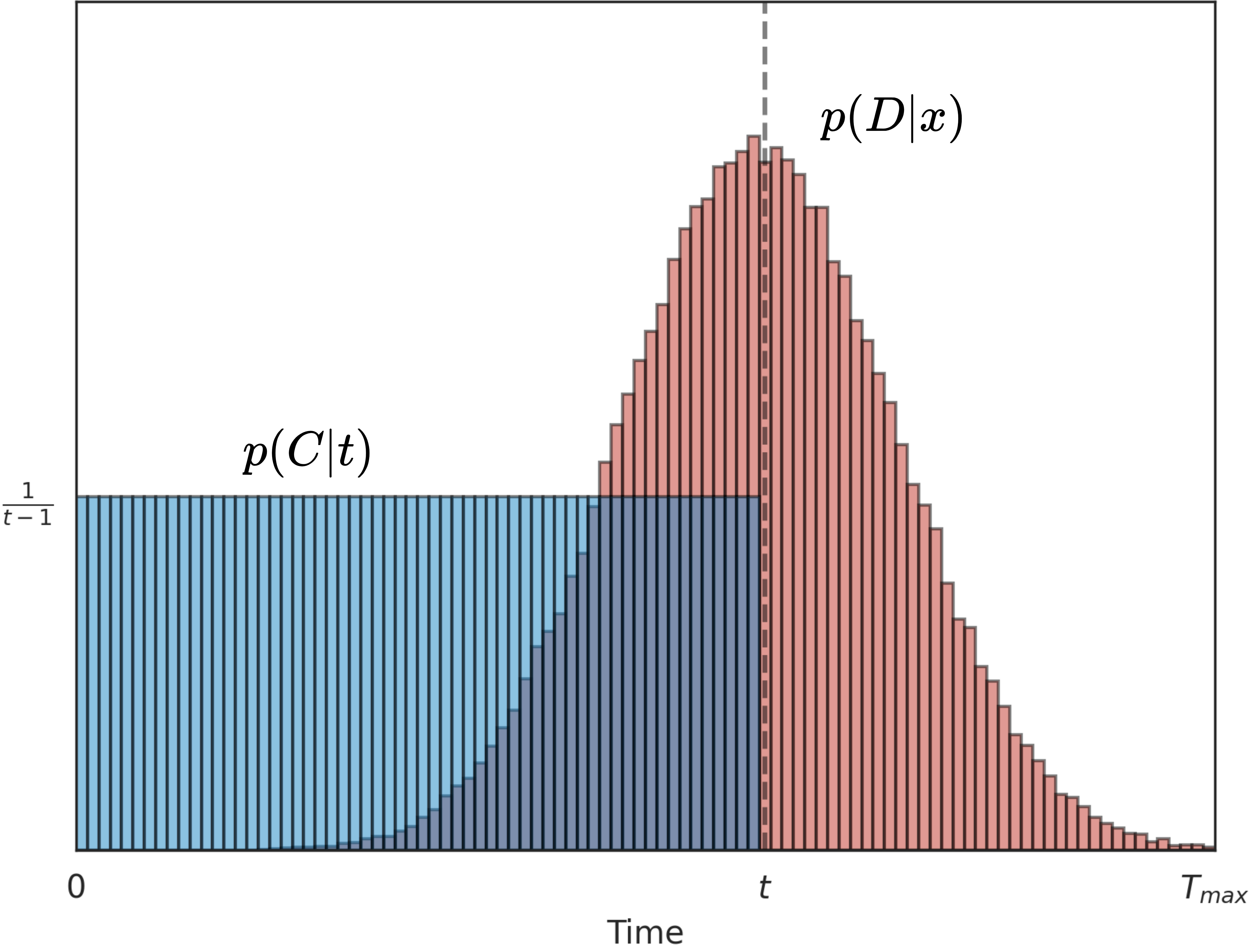}
        \caption{\method data generation mechanism.}
        \label{fig:ours}
    \end{subfigure}
    \hspace{0.025\textwidth}
    \begin{subfigure}{0.45\textwidth}
        \centering
        \includegraphics[width=0.9\textwidth]{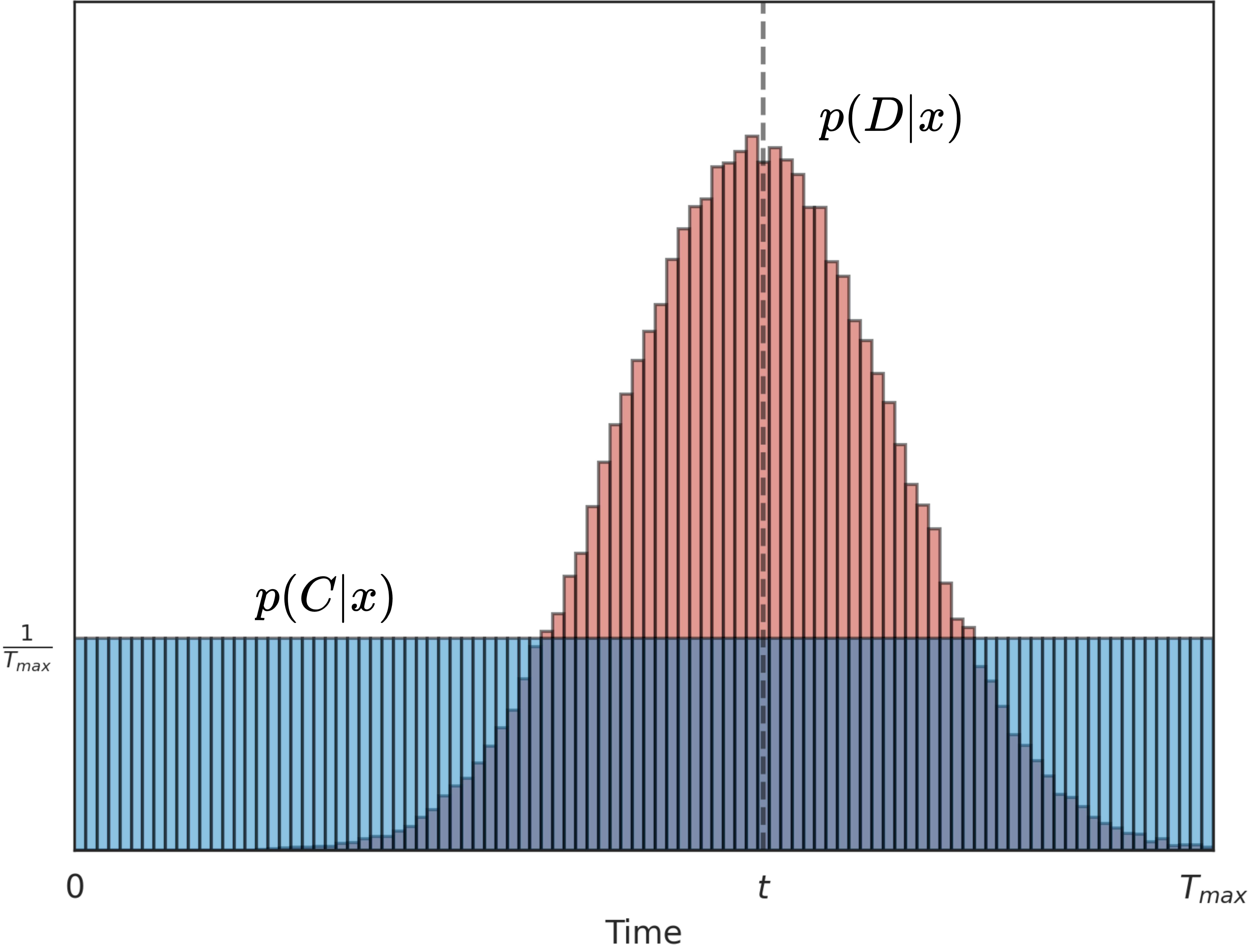}  
        \caption{The classical data generation mechanism.}
        \label{fig:classical}
    \end{subfigure}
    \hspace{0.025\textwidth}
    \caption{Distributional survival analysis data generation mechanisms. (a) In the proposed event-conditional censoring model (\method), $t$ is drawn from the death time distribution and $c$ is uniformly sampled up to $t$. (b) In the classical model, $t$ and $c$ represent randomly drawn death and censoring times from the corresponding distributions. If $c < t$, the patient is censored and the observation is the censoring time. Otherwise, the patient is uncensored and observation is the death time.}
    \label{fig:datageneration}
\end{figure*}

We still need to make two assumptions -- the censoring distribution $p(C|D,x)$ and the event distribution $p_\theta(D=t|x)$. We define the event distribution $p_\theta(D=t|x)$ below in \secref{sec:death_dist} and here we define the censoring distribution $p(C|D,x)$.  In principle, this can also be learned from the data but for simplicity we assume a uniform censoring distribution $p(C=c|D=t,x)= const$ for $c<t$ and 0 elsewhere (see \figref{fig:ours}), giving 
\beq
p_\theta(C=c|x) = \sum_{t=c+1}^{\tmax} \frac{1}{t-1} p_\theta(D=t|x)
\eeq
For any event distribution model $p_\theta(D=t|x)$ the likelihood objective to maximise is
\beq
    \mathbb{L}(\theta) 
    \equiv \sum_{n\in \uncensSet} \log{p_\theta(D=t_n|x_n)} + \sum_{i\in \censSet} {\log{\sum_{t=c_i+1}^{\tmax} \frac{1}{t-1}p_\theta(D=t|x_i)}} \numberthis
    \label{eq:our:model}
\eeq
\subsection{Event time distribution}
\label{sec:death_dist}
We need to make an appropriate choice for the event time distribution $p_\theta(D=t|x)$. We employ a discretised form of the Gaussian distribution
\beq
p_{\theta}(D=t|x) = \frac{1}{Z} \exp\br{\frac{-(t-\mu_{\theta}(x))^2}{2\sigma^2_{\theta}(x)}}
\label{eq:disc_gauss}
\eeq
In this formulation, $\mu_{\theta}(x)$ and $\sigma_{\theta}(x)$ are parameters of the distribution that are predicted by the model (a neural network parameterized by $\theta$), and $Z$ is a normalisation factor, defined as
\beq
Z = \sum_{t=1}^{\tmax} \exp\br{\frac{-(t-\mu_{\theta}(x))^2} {2\sigma^2_{\theta}(x)}}
\eeq
This formulation has the following advantages
\begin{itemize}
    \item The term $(t-\mu_\theta(x))^2$ ensures a heavier penalty for predictions that deviate significantly from the true death time, promoting closer predictions. This stands in contrast to approaches that treat death times as independent categories \cite{lee2018}, which do not fully capture this relationship.
    \item The model only outputs two quantities $(\mu_\theta(x), \sigma_\theta(x))$. This keeps the number of parameters low, reducing the risks of overfitting compared to treating this as a $T_{max}$ classification task, with category for each timepoint \cite{lee2018}.
\end{itemize}
In principle, the form of the distribution $p_{\theta}(D=t|x)$ is also learnable, but we found that the discrete Gaussian performed well in our experiments.
\section{Previous works}
\subsection{Classical censoring model}
\label{sec:classical}

One common approach in the literature is to assume that censoring times follow a distribution $p(C=c|x)$ and death times follow a distribution $p_\theta(D=t|x)$. These times are independently sampled and then compared: if the censoring time is less than the death time, the observation is the censoring time; otherwise, it is the death time \cite{lee2018,klein2003}, see \figref{fig:classical}. This leads to the following model
\beq
p_\theta(\delta,c,t|x) = p_\theta(t|x)p(c|x)p(\delta|c,t)
\eeq
where $p(\delta=1|c,t)=1$ if $c\geq t$ and $p(\delta=0|c,t)=1$ if $c<t$. For a uniform censoring distribution $p(C=c|x) = \frac{1}{\tmax}$ a (right) censored observation then has the following likelihood
\beq
p_\theta(\delta=0,C=c|x) = \frac{1}{\tmax}\sum_{t=c+1}^{\tmax} p_\theta(D=t|x)
\eeq
and the likelihood of an uncensored observation is given by
\beq
    p(\delta=1, D=t|x) = \frac{\tmax-t+1}{\tmax} p_\theta(D=t|x)
\eeq
Omitting additive constants, the objective then is to maximize
\beq
    \mathbb{L}(\theta) \equiv \sum_{n\in \uncensSet} \log{p_\theta(D=t_n|x_n)} + \sum_{i\in \censSet} {\log{\sum_{t=c_i+1}^{\tmax} p_\theta(D=t|x_i)}}
    \label{eq:classical}
\eeq
Comparing this with the \method censoring mechanism (\eqref{eq:our:model}), the difference is the $1/(t-1)$ factor in the censored summation term. Also, for this classical approach, if one wanted to generate data from the model, one cannot a priori decide on how many samples are censored or uncensored. The generation process in \eqref{eq:classical} generates either a censored or uncensored datapoint, with the probability of this happening being a function of $\theta$. 

\subsection{Cox Model\label{sec:cox}}
We briefly review the standard Cox proportional hazards approach \cite{cox72}.  The Cox model is ubiquitous in survival analysis -- however, it cannot directly predict the death time nor deal easily with censored data. Both of these issues are, we believe, vital for modern survival analysis applications.  Whilst the Cox model does not directly produce a prediction for the death time, there are standard approaches to estimate the death time \cite{breslow1974}, and as such it is an important baseline comparison method. The hazard function $h(t)$ models the chance that a patient will die in an infinitesimal time interval $[t, t + \Delta{}t)$ given that death has not occurred before
\beq
h(t)=\lim_{\Delta t \to 0}\frac{p(D \in[t, t+\Delta t)|D \geq t)}{\Delta t}
\eeq
The \CoxPH model \cite{cox72} constrains the hazard function (conditioned on the patient covariates $x$) to the form
\beq
    h(t|x)=h_0(t) \exp(g_\theta(x))
\eeq
Here $h_0(t)$ is the baseline hazard function, which depends only on $t$, while $g_\theta(x)$ depends on the patient covariates $x$ and $\theta$ are the model parameters. The standard \CoxPH model \cite{cox72} further constrains the hazard function (conditioned on the patient covariates $x$) to the linear form $g_\theta(x) = \beta x$. DeepSurv \cite{katzman2018} and other deep neural networks extend it to non-linear $g_\theta(x)$. For each patient $n$ we define the risk set $R_n$ as all those patients that have not died before patient $n$ and define the relative death risk as
\beq
    p(D_n=t_n|R_n)  = \frac{h(t_n|x_n)}{\sum_{m \in R_n} h(t_m|x_m)} = \frac{\exp(g_\theta(x_n))}{\sum_{m \in R_n}\exp(g_\theta(x_m))}
    \label{eqn:hazard}
\eeq
The partial log-likelihood is then defined as the sum of $\log p(D_n=t_n|R_n)$ for the set of uncensored patients $\uncensSet$
\beq
L(\theta) \equiv \frac{1}{|\uncensSet|} \sum_{n \in {\uncensSet}}  \sq{g_\theta(x_n) - \log{\sum_{m \in R_n}\exp(g_\theta(x_m))}}
\label{eq:cox}
\eeq
As can be seen from \eqref{eq:cox}, Cox-based methods utilize the censored data only in constructing the risk set $R_n$ and maximise the likelihood that the uncensored patients die before patients in the risk set. In our experiments, $x$ is a high-dimensional CT scan, and the function $g_\theta$ is a deep neural network that is costly to compute in both time and memory. A typical approach to optimising \eqref{eq:cox} is stochastic gradient descent, which involves selecting minibatches of training observations at each iteration $m^i \in \dataSet$, where $m^i$ denotes the minibatch index set at iteration $i$ \cite{katzman2018}. The objective then is to maximise
\beq
L(\theta^i) \equiv \frac{1}{|\uncensSet^i|} \sum_{n \in {\uncensSet^i}}  \sq{g_{\theta^i}(x_n) - \log{\sum_{m \in R_n^i}\exp(g_{\theta^i}(x_m))}}
\label{eq:cox:sgd}
\eeq
where $\theta^i$ represents the model parameters at iteration $i$, $\uncensSet^i$ is the uncensored observation index set for samples in minibatch $m^i$, and $R_n^i$ is the risk set for patient $n$ in the same minibatch.

However, \eqref{eq:cox:sgd} is a ranking objective that compares patients within the minibatch based on their predicted mortality risk. This requires large minibatch sizes for robust training; however, for high-resolution input (\eg 3D CT scans), we are limited by GPU memory to small minibatch sizes. Consequently, the minibatches often contain only censored patients, \ie $\uncensSet^i= \emptyset$. In such cases, \eqref{eq:cox:sgd} is undefined, and these minibatches are excluded from the training process, resulting in a significant reduction in the training data. To overcome this, we use a memory bank \cite{wu2018,he2016} to store neural network predictions for later iterations \cite{shahin22}, see \appref{app:coxmb} for details. We call this approach CoxMB and compare it with the standard Cox model in our experiments.

\subsection{DeepSurv}
DeepSurv is a deep neural network that is trained using the Cox objective function (\eqref{eq:cox}), outputting a single scalar value that represents the risk of death \cite{katzman2018}. It is compared with our CoxMB model, which uses a memory bank to store the risk of death for each patient during training, using this information to penalise the model for inaccuracies in predicting the ranking of patients' survival times.

\subsection{DeepHit}
\citet{lee2018} approach survival analysis as a classification task with $\tmax$ categories. Specifically, a neural network predicts a vector of $\tmax$ values, which a softmax function then transforms into a death distribution, $p_\theta(D=t|x)$. This approach, however, has a few challenges: (1) the ordinal nature of the death time is not directly captured because the softmax function regards different death times as separate classes; (2) if $\tmax$ is large, the model requires more parameters, heightening the risk of overfitting; (3) some death times might not be represented in the training data, which could reduce softmax probabilities to zero, yielding no gradient and impeding the learning process for these times. All of these issues are addressed by our alternative formulation in \secref{sec:death_dist}. 

To leverage the censored data, DeepHit uses a combination of the classical censoring model \eqref{eq:classical} and a ranking objective. Specifically, the objective function is composed of two terms $\mathcal{L}_{\text{DeepHit}} = \mathcal{L}_\text{lik.}^c + \mathcal{L}_{\text{rank.}}$, where $\mathcal{L}_\text{lik.}^c$ represents the classical likelihood (\eqref{eq:classical}) with a softmax function to model the death time distribution, and $\mathcal{L}_{\text{rank.}}$ is a ranking term that penalises the model for inaccuracies in predicting the ranking of patients' survival times, mirroring the Cox objective
\beq
\mathcal{L}_{\text{rank.}} = \eta(F_\theta(t_i|x_i), F_\theta(t_i|x_j)) \quad \forall i,j \in \dataSet \quad \text{s.t.} \quad t_i < t_j
\eeq
where $\eta(x,y)=\exp(\frac{-(x-y)}{s})$, with $s$ being a hyperparameter set to 0.1, following the official implementation. $F_\theta(t|x)$ represents the cumulative distribution function of the predicted distribution $p_\theta(t|x)$. 

\subsection{DeepHit ($\mathcal{L}_\text{lik.}^c$ only)}
To evaluate the contribution of the likelihood term in DeepHit, we train another model with the same architecture but without the ranking term $\mathcal{L}_{\text{rank.}}$.

\section{Experiments}
We evaluate our methods on a practical and challenging real data problem. IPF is a chronic fibrotic interstitial lung disease of unknown cause, associated with progressive fibrosis (stiffening and scarring of lung tissue), deterioration of lung function, and shortened survival \cite{lederer2018idiopathic,barratt2018idiopathic}. Survival analysis of patients with IPF is fundamental for studies that evaluate factors associated with disease progression and is part of the analysis of clinical drug trials. However, it is a challenging task due to the heterogeneous progression trajectories of IPF and the lack of available mortality predictors and survival models. Cox models are often used in these studies to identify associations with mortality \cite{jacob2017,gao2021a}.  Despite its popularity, the Cox model has several limitations. Primarily, it relies on the assumption of proportional hazards, which states that the relative hazard remains constant over time between different patients. This assumption is not always accurate, particularly in progressive diseases such as IPF. Furthermore, the Cox model estimates the relative hazard, rather than the actual death time, which is often more useful and easier to interpret.

More similar to our method, other approaches train models to predict death time, rather than ranking patients according to their death risk. One notable example of this approach is DeepHit \cite{lee2018}, which uses a fully connected layer in a deep network to output the probability of death at every possible time. This approach treats death-time prediction as a one-of-$\tmax$ classification problem and does not encode the natural assumption that making a small error in the time of death should be penalised less than predicting a large error in the time of death.

\subsection{Dataset and preprocessing}
We use the Open Source Imaging Consortium (OSIC)\footnote{\href{https://www.osicild.org/dr-about.html}{https://www.osicild.org/dr-about.html}} dataset which encompasses lung CT scans along with contemporaneous clinical data in addition to mortality labels in months ($\delta$ and $t$ if $\delta=1$, otherwise $c$). We examine the performance of different methods using exclusively CT images or a combination of CT images and clinical data, as each contains pertinent information related to disease progression in IPF. The dataset consists of 728 samples, which we randomly divided into training (70\%), validation (15\%), and test (15\%) sets. The mean and standard deviation of the metrics are reported over five runs with different random splits. Approximately 65\% (470 samples) of the dataset are right-censored.

For the imaging data, only CT scans with a slice thickness of $\leq$ 3mm are considered. All scans are cropped to the lung area using the lung segmentation model trained by \cite{Hofmanninger2020}. These scans are then resampled to achieve an isotropic pixel spacing of $1\times1\times1$ mm$^3$ via linear interpolation. Following this, the scans are resized to dimensions of $256\times256\times256$ voxels using bicubic interpolation. Later, we apply histogram matching and a windowing operation within the range [-1024, 150] Hounsfield Units to remove irrelevant information. Finally, we normalize the scans to have zero mean and unit variance based on the statistics drawn from the training set. We apply random rotation (up to 15 degrees) and translation (up to 20 pixels) to augment the training data.

In experiments involving clinical data, we incorporate six clinical features: age, sex, smoking history (categorised as never-smoked, ex-smoker, or current smoker), antifibrotic treatment (yes or no), Forced Vital Capacity (FVC) percent, and carbon monoxide diffusion capacity (DLCO). To ensure the correspondence between the imaging and clinical data, we only include patients whose lung function tests were performed within 3 months of the CT scan. Continuous features (age, FVC percent and DLCO) are normalised to have zero mean and unit variance, while categorical features are transformed via one-hot encoding. Missing values are sampled using a latent variable model following \cite{shahin22}. During testing, we use the most probable value from the missing data imputation model.
\subsection{Implementation details}
In our experimental setup, the event distribution models parameterize the distribution $p_{\theta}(t|x)$ using $\mu_{\theta}$ and $\sigma_{\theta}$. A deep learning model parameterized by $\theta$ is used to learn $\mu_{\theta}$, while $\sigma$ is fixed at 12 months. This helps to stabilise the training process and mitigate overfitting (see \cite{nix1994} for a similar observation). For DeepHit, the output of the model is a vector of size $\tmax$, representing the logits of the 1-of-$\tmax$ classification labels. Finally, the DeepSurv and CoxMB models output a single scalar that represents the predicted risk of death, $g_\theta(x)$ in \eqref{eq:cox}. We evaluate the performance of the models when trained on imaging data exclusively, as well as combined imaging and clinical data.

To process HRCT scans, we use a 3D Convolutional Neural Network (CNN), as illustrated in \appref{app:model_architecture} (left). The network initiates with a 3D convolutional layer, which is followed by an instance normalization layer and a leaky ReLU activation function. We then stack four residual blocks, each comprising three 3D convolutional layers \cite{he2016}. After each convolutional layer, we use instance normalisation \cite{ulyanov2016} and leaky ReLU \cite{maas2013} layers. We utilized $1\times1\times1$ kernels for the first and last convolutional layers, while the middle layer used a $3\times3\times3$ kernel. In a parallel branch, we use a single convolutional layer, and the outputs of the two branches are concatenated. The output of this series of layers is then passed through another convolutional layer, designed with a stride of 2, to halve the spatial dimension. Finally, we use a convolutional layer with 16 filters and a $1\times1\times1$ kernel to produce a compact feature representation. We flatten this representation and input it into the final fully connected layer. In designing this network, we were aware that the progression of IPF manifests itself in fine pulmonary patterns, such as honeycombing, reticulation, and ground glass opacities. To capture these nuances, we opt for small kernels and deliberately avoid pooling layers, as this could result in the loss of fine image details.

When we incorporate clinical data, we use a Multi-Layer Perceptron (MLP) that consists of two fully connected layers with 32 neurones each, each followed by batch normalisation \cite{ioffe2015} and leaky ReLU activation \cite{maas2013}, as detailed in \appref{app:model_architecture} (right). The MLP output is concatenated with the CNN output. The CNN output, which represents imaging data, is projected to a 32-element vector to balance the contributions from both imaging and clinical data. The combined output is subsequently propagated through a final fully connected layer.

For optimisation, we use AdamW optimiser \cite{loshchilov18} with a learning rate of $10^{-4}$ for the classical and event-conditional censoring models and $5\times10^{-4}$ for DeepHit, DeepSurv, and CoxMB. The optimal learning rate value was tuned via a random search based on the performance on the validation set. Additionally, we apply a cosine annealing learning rate scheduler and gradient clipping. Due to the high resolution of the imaging data (256x256x256), we use a batch size of 2 for all models. We train the models for an initial 300 epochs. However, training is halted if there is no improvement in validation performance for 50 consecutive epochs. In CoxMB, we use a $K$ value of 1.0. We use a $\tmax$ of 156 months for all models, which is the maximum observed time in the dataset. The models are implemented using PyTorch and trained on a single NVIDIA A6000 GPU.

\subsection{Evaluation metrics}
\paragraph{Concordance index}
The $\text{C-Index}$ estimates the probability that the predicted risks or survival times of a randomly chosen pair of patients will have the same ordering as their actual survival times \cite{Harrell96}
\beq
\text{C-Index} = \frac{\# \text{concordant pairs}}{\# \text{concordant pairs} + \# \text{discordant pairs}}
\eeq
A pair is considered concordant if the ranking predicted by the model matches the true ranking, and discordant if it does not. A perfect model will have a $\text{C-Index}=1$. It is worth noting that the C-Index is a ranking metric, which only assesses the order in which the predicted values should be ranked compared to the true ranking.

\paragraph{Mean Absolute Error} The MAE assesses the difference between death times predicted by the model and the true death times
\beq
\text{MAE} = \frac{1}{|\uncensSet|} \sum_{i\in \uncensSet} |\hat{t}_i - t_i|
\eeq
where $\hat{t}_i$ is the predicted death time for patient $i$.

\paragraph{Relative Absolute Error} We also report the RAE which quantifies the relative deviation of the predicted time from the true death time
\begin{align}
    \text{RAE} &= \frac{1}{|\uncensSet|} \sum_{i\in\uncensSet} \frac{|\hat{t}_i-t_i|}{t_i}
\end{align}
\begin{table*}[!t]
    \centering
    \caption{Comparison of the test performance of the different methods on OSIC dataset when trained on imaging data only, as well as combined imaging and clinical data. The mean and standard deviation are reported over five runs with different random train/val/test splits. The best results are highlighted in bold.}
    \label{tab:main_results}
    \begin{tabular}{@{} m{0.5cm} m{4.3cm} r@{\text{} $\pm$ \text{}}l r@{\text{} $\pm$ \text{}}l r@{\text{} $\pm$ \text{}}l @{}}
    \toprule
    Data & Method & \multicolumn{2}{c}{C-Index $\uparrow$} & \multicolumn{2}{c}{MAE $ \downarrow$} & \multicolumn{2}{c}{RAE $\downarrow$} \\ \midrule
    \multirow{6}{*}{\rotatebox[origin=c]{90}{Imaging}} & DeepSurv (Cox) & 67.441 & 4.572 & 44.898 & 19.505 & 2.286 & 1.414\\
    & CoxMB & \textbf{71.067} & \textbf{5.572} & 28.887 & 2.315 & 1.762 & 0.807\\
    & DeepHit & 53.165 & 8.313 & 31.074 & 7.765 & 1.830 & 0.522\\
    & DeepHit ($\mathcal{L}_\text{lik.}^c$ only) & 57.607 & 4.813 & 29.862 & 3.742 & 1.926 & 0.869\\
    & Classical Censoring & 68.844 & 5.313 & 20.448 & 4.787 & 1.407 & 0.853\\
    & \method & 69.273 & 0.946 & \textbf{19.319} & \textbf{1.613} & \textbf{1.338} & \textbf{0.665} \\
    \midrule
    \multirow{6}{*}{\rotatebox[origin=c]{90}{\parbox{1.5cm}{Imaging + Clinical}}} & DeepSurv (Cox) & \textbf{72.1} & \textbf{2.186} & 27.603 & 3.345 & 1.718 & 0.742\\
    & CoxMB & 68.877 & 2.413 & 24.413 & 2.548 & 1.892 & 0.868\\
    & DeepHit & 54.980 & 3.490 & 31.246 & 4.599 & 2.240 & 0.862\\
    & DeepHit ($\mathcal{L}_\text{lik.}^c$ only) & 52.882 & 3.843 & 28.718 & 2.077 & 2.059 & 0.722\\
    & Classical Censoring& 70.35 & 2.947 & 20.476 & 1.85 & 1.546 & 0.611 \\
    & \method & 70.957 & 3.048 & \textbf{19.178} & \textbf{0.795} & \textbf{1.48} & \textbf{0.671} \\
    \bottomrule
    \end{tabular}
\end{table*}
\subsection{Results}
The evaluation of survival analysis performance depends on the particular clinical objective. For instance, if the aim is to stratify patients into high and low-risk groups, the C-Index is a suitable metric. In contrast, if the objective is a precise prediction of the time of death for each patient, metrics such as MAE and RAE are more appropriate.

In \tabref{tab:main_results}, we report the test performance of the different methods on the OSIC dataset. For the Cox-based methods, we notice that the introduction of memory banks during training (CoxMB) leads to a significant performance improvement compared to the DeepSurv model, which employs the standard Cox objective function \cite{cox72,katzman2018}. This improvement can be seen through the increase in C-Index by 3.63, a reduction of the MAE by 16 months, and a decrease in the RAE by 0.046.

Upon inclusion of clinical data, CoxMB upholds superior performance on MAE in contrast to DeepSurv, whereas DeepSurv excels in ranking performance. This performance divergence, particularly with respect to the decline of the C-Index in the CoxMB case, can likely be attributed to the high noise level and the presence of missing values in clinical data. In general, DeepSurv seems to benefit more from the inclusion of clinical data than CoxMB, where the improvements are marginal. CoxMB already performs well on the imaging data, and the clinical data do not provide much additional information.

For distribution-based methods, \method outperforms all other distribution-based baselines in C-Index, MAE, and RAE metrics,
whether trained solely on imaging data or a combination of imaging and clinical data. 
The superiority of our method is particularly noticeable in the hybrid case, where the the MAE decreases by 9.92 and 1.3 months compared to the DeepHit and the classical censoring models, respectively. Similarly, the C-Index improves by 12.22 and 0.61 compared to these models. Comapred to DeepSurv and CoxMB, \method offers a remarkable improvement in MAE (8.43 and 5.23 months, respectively) and a comparable ranking performance. This demonstrates the effectiveness of \method in efficiently capturing the censoring process. Interestingly, \method significantly outperforms DeepHit. In addition to the different modelling of the censoring process, this can be attributed to the different ways each model handles the event distribution. \method applies a discretized version of the Gaussian distribution (as per \eqref{eq:disc_gauss}), whereas DeepHit considers it as a classification problem comprising $\tmax$ classes, executed using a fully-connected layer followed by a softmax function. By disregarding the ordinal nature of the time variable and facing the potentially large class number, $\tmax$, DeepHit is more susceptible to overfitting.

In summary, \method outperforms all the baselines in predicting the time of death for IPF patients, whether trained solely on imaging data or a combination of imaging and clinical data. Additionally, it delivers competitive C-Index performance despite not incorporating a ranking objective. This makes it a more appropriate choice for clinical scenarios where the precise prediction of the time of death takes precedence over the ranking of patients' survival times. On the other hand, if the ranking of survival times is of paramount importance, CoxMB model offers a more robust training strategy by employing memory banks, especially beneficial when training on high-resolution imaging data.
\begin{figure*}[!t]
    \centering
    \includegraphics[width=1.0\textwidth]{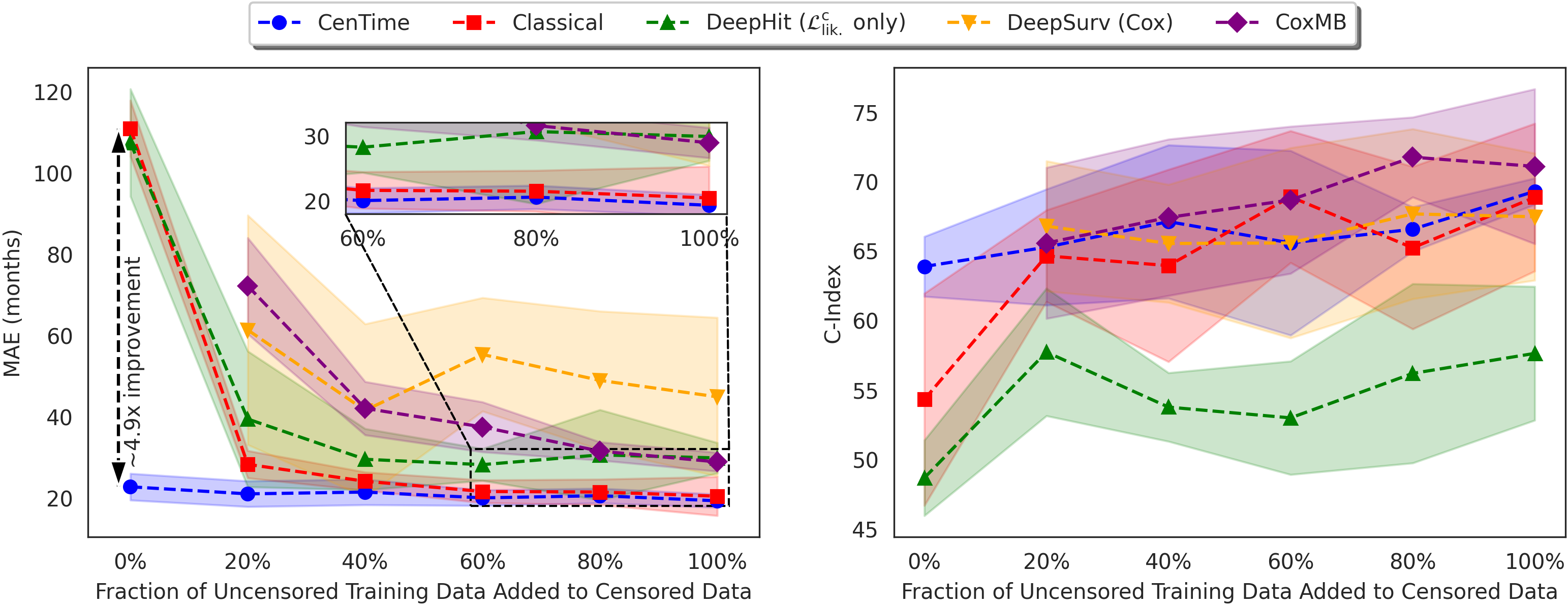}
    \caption{Performance of the different methods when trained on gradually increasing percentages of uncensored data added to the censored data. $0\%$ corresponds to training on purely censored data, while $100\%$ corresponds to training on the full training set. The mean and standard deviation are reported over five runs with different random train/val/test splits.
    }
    \label{fig:limited_uncensored_data}
\end{figure*}
\subsubsection{Performance under limited uncensored training data}
\label{sec:limited_data}
The amount of uncensored data available for training survival models is typically limited. Therefore, it is critical for learning algorithms to use the available censored data effectively to improve performance. In this subsection, we examine the performance of the different methods when trained on a limited amount of uncensored data, in addition to the censored data (imaging only). We randomly sample 0\% (purely censored), 20\%, 40\%, 60\%, 80\%, and 100\% of the uncensored data. In each scenario, all the censored data is added to compose the training set. The results are presented in \figref{fig:limited_uncensored_data}.

The initial observation is that Cox-based models (DeepSurv and CoxMB) are only trainable when uncensored examples are available during training. This is because the objective function is defined solely for uncensored examples (see \eqref{eq:cox}). Second, when utilising purely censored data, \method shows a significant improvement ($\approx 4.9$x in terms of MAE) over the classical and DeepHit models. This is because \method forms a consistent estimator of the model parameter $\theta$ even with purely censored data, a feature not shared by the classical and DeepHit models. As the amount of uncensored data included in the training data increases, we generally observe an improvement in the performance of all models, and the differences between the various methods diminish. However, \method continues to outperform the other methods in terms of MAE and offers competitive performance in terms of the C-Index. These findings underscore the effectiveness of our proposed approach in modelling the censoring process and utilizing it efficiently.

Furthermore, we observe that the performance of the CoxMB model, when trained with a limited amount of uncensored data, is comparable to that of the DeepSurv model. This can be attributed to the lessened effectiveness of the memory bank when the amount of uncensored data is limited. However, as the amount of uncensored data increases, the memory bank efficacy improves and the performance of CoxMB consistently surpasses that of the DeepSurv model. This is evident in both the C-Index and the MAE metrics. Intriguingly, the C-Index performance of \method is comparable to that of DeepSurv, despite the fact that it does not use a ranking objective. This further underlines the robustness and versatility of our proposed event-conditional censoring model.

\subsubsection{Effect of lung segmentation}
\begin{figure}[t]
    \centering
    \includegraphics[width=0.4\columnwidth]{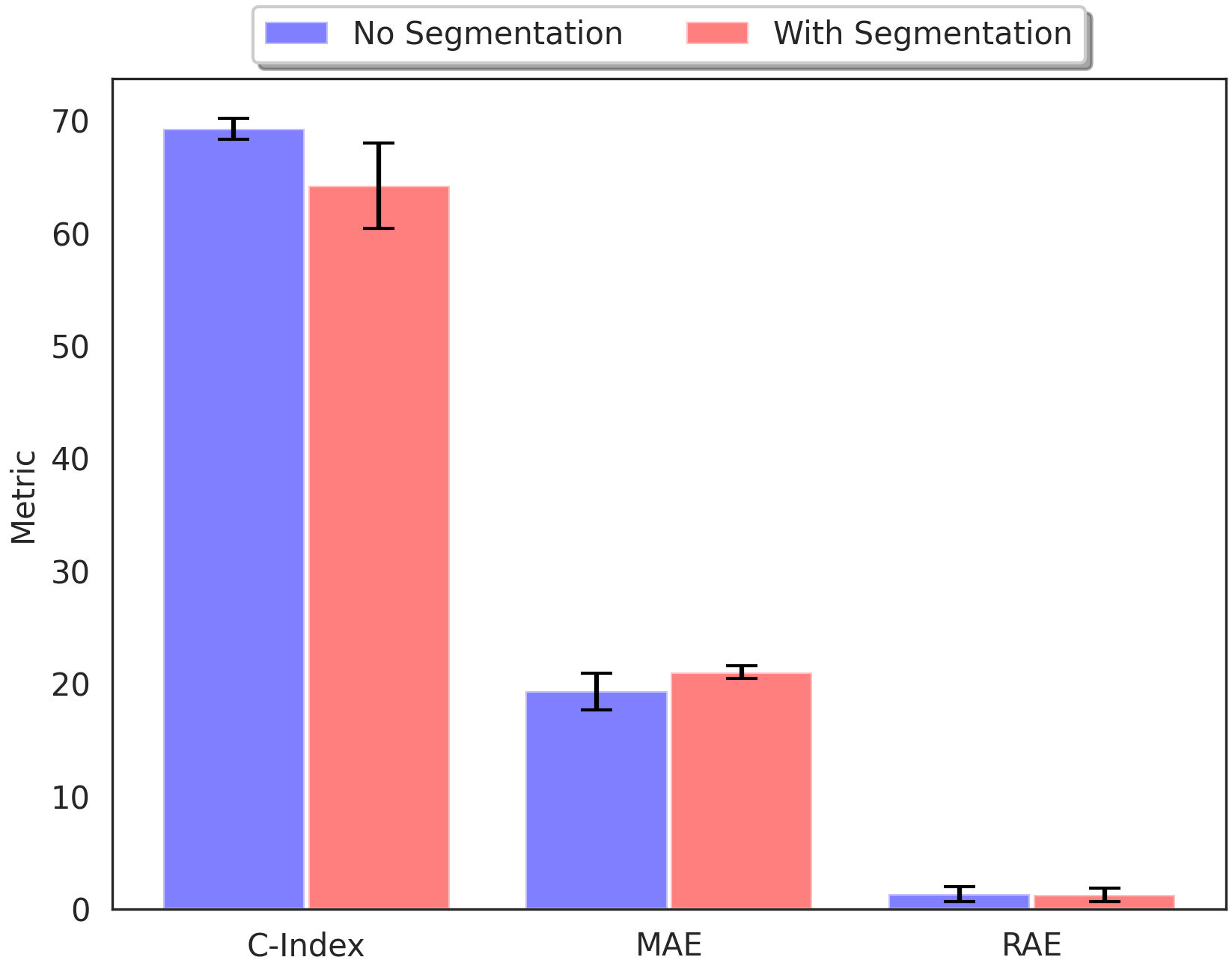}
    \caption{Effect of lung segmentation on the performance of \method.}
    \label{fig:lung_segmentation}
\end{figure}

Idiopathic Pulmonary Fibrosis predominantly affects the lungs, making this area the most relevant in CT scans. However, there is some evidence suggesting that the disease can also affect other organs, such as the heart \cite{agrawal2016}. Therefore, we examine the effect of lung segmentation on the performance of \method, when trained on imaging data. We train the model with and without lung segmentation (using \cite{Hofmanninger2020}) and report the results in \figref{fig:lung_segmentation}. We do not observe a significant difference in the performance, which suggests that the model is able to learn the relevant features from the lung area without the need for explicit segmentation. This also allows the model to benefit from information in the non-lung area (\eg heart) if it is relevant to the survival prediction task.

\section{Conclusions}
Our work demonstrates the limitations of existing survival methods and addresses them. Traditional Cox-based methods \textbf{(i)} assume the strong proportional hazards assumption, which is not always true, \textbf{(ii)} estimate the relative hazard rather than the actual death time, which is often more useful and easier to interpret, and \textbf{(iii)} represent a ranking method and, therefore, require a large batch size, which is not always feasible. DeepHit \textbf{(iv)} does not encode the ordinal nature of the target survival time variable, \textbf{(v)} approaches the problem as a classification task, which becomes prone to overfitting with a large number of classes. Our \method model addresses all these limitations. By modelling the death and censoring likelihoods, it circumvents the hazards proportionality assumption \textbf{(i)}, directly estimates the death time \textbf{(ii)}, and imposes no batch size restrictions \textbf{(iii)}. Furthermore, because of the adoption of the discretised Gaussian distribution, our model naturally encodes the ordinal nature of the target survival time variable \textbf{(iv)} and, by outputting only the discretised Gaussian distribution parameters, is less susceptible to overfitting \textbf{(v)}. Finally, compared to the classical censoring mechanism, \method offers a convenient alternative to the classical censoring model by providing a consistent estimator even with purely censored data alone and should be particularly useful in situations with only very limited uncensored entries.

Our results underscore the effectiveness of \method in predicting the time of death, while also offering competitive performance in terms of ranking, even without a ranking objective. This makes \method a compelling choice for clinical scenarios where accurate prediction of death time takes precedence over the ranking of patients' survival times, particularly when dealing with limited observed death time data.
\section*{Acknowledgments}
This work was supported by the Open Source Imaging Consortium (OSIC) \href{https://www.osicild.org}{osicild.org}.
\bibliographystyle{unsrtnat}
\bibliography{references}

\begin{thebibliography}{30}
\providecommand{\natexlab}[1]{#1}
\providecommand{\url}[1]{\texttt{#1}}
\expandafter\ifx\csname urlstyle\endcsname\relax
  \providecommand{\doi}[1]{doi: #1}\else
  \providecommand{\doi}{doi: \begingroup \urlstyle{rm}\Url}\fi

\bibitem[Lee and Lim(2019)]{lee2019}
Seungyeoun Lee and Heeju Lim.
\newblock {Review of statistical methods for survival analysis using genomic
  data}.
\newblock \emph{Genomics \& informatics}, 17\penalty0 (4), 2019.

\bibitem[Lee et~al.(2018)Lee, Zame, Yoon, and Van Der~Schaar]{lee2018}
Changhee Lee, William Zame, Jinsung Yoon, and Mihaela Van Der~Schaar.
\newblock {Deephit: A deep learning approach to survival analysis with
  competing risks}.
\newblock In \emph{{Proceedings of the AAAI conference on artificial
  intelligence}}, volume~32, 2018.

\bibitem[Zhao et~al.(2022)Zhao, Shahin, Zhou, Gudmundsson, Szmul, Mogulkoc, van
  Beek, Brereton, van Es, Pontoppidan, Savas, Wallis, Unat, Veltkamp, Jones,
  van Moorsel, Barber, Jacob, and Alexander]{zhao2022prognostic}
An~Zhao, Ahmed~H. Shahin, Yukun Zhou, Eyjolfur Gudmundsson, Adam Szmul, Nesrin
  Mogulkoc, Frouke van Beek, Christopher~J. Brereton, Hendrik~W. van Es,
  Katarina Pontoppidan, Recep Savas, Timothy Wallis, Omer Unat, Marcel
  Veltkamp, Mark~G. Jones, Coline H.~M. van Moorsel, David Barber, Joseph
  Jacob, and Daniel~C. Alexander.
\newblock {Prognostic Imaging Biomarker Discovery in Survival Analysis for
  Idiopathic Pulmonary Fibrosis}.
\newblock In \emph{{International Conference on Medical Image Computing and
  Computer-Assisted Intervention}}, pages 223--233, 2022.

\bibitem[Shahin et~al.(2022)Shahin, Jacob, Alexander, and Barber]{shahin22}
Ahmed~H. Shahin, Joseph Jacob, Daniel Alexander, and David Barber.
\newblock {Survival Analysis for Idiopathic Pulmonary Fibrosis using CT Images
  and Incomplete Clinical Data}.
\newblock In \emph{Proceedings of The 5th International Conference on Medical
  Imaging with Deep Learning}, volume 172 of \emph{Proceedings of Machine
  Learning Research}, pages 1057--1074, 2022.

\bibitem[Lu et~al.(2023)Lu, Aslani, Zhao, Shahin, Barber, Emberton, Alexander,
  and Jacob]{lu2023}
Yaozhi Lu, Shahab Aslani, An~Zhao, Ahmed~H. Shahin, David Barber, Mark
  Emberton, Daniel~C Alexander, and Joseph Jacob.
\newblock {A hybrid CNN-RNN approach for survival analysis in a Lung Cancer
  Screening study}.
\newblock \emph{arXiv preprint arXiv:2303.10789}, 2023.

\bibitem[Richardeau and Pham(2012)]{richardeau2012}
Frederic Richardeau and Thi Thuy~Linh Pham.
\newblock Reliability calculation of multilevel converters: Theory and
  applications.
\newblock \emph{IEEE Transactions on Industrial Electronics}, 60\penalty0
  (10):\penalty0 4225--4233, 2012.

\bibitem[Kim and Suk~Kim(2014)]{kim2014}
Juyoung Kim and Myung Suk~Kim.
\newblock Analysis of automobile repeat-purchase behaviour on crm.
\newblock \emph{Industrial Management \& Data Systems}, 114\penalty0
  (7):\penalty0 994--1006, 2014.

\bibitem[Emmert-Streib and Dehmer(2019)]{emmert2019introduction}
Frank Emmert-Streib and Matthias Dehmer.
\newblock {Introduction to survival analysis in practice}.
\newblock \emph{Machine Learning and Knowledge Extraction}, 1\penalty0
  (3):\penalty0 1013--1038, 2019.

\bibitem[Wang et~al.(2019)Wang, Li, and Reddy]{wang2019machine}
Ping Wang, Yan Li, and Chandan~K Reddy.
\newblock {Machine learning for survival analysis: A survey}.
\newblock \emph{ACM Computing Surveys (CSUR)}, 51\penalty0 (6):\penalty0 1--36,
  2019.

\bibitem[Kaplan and Meier(1958)]{kaplan1958nonparametric}
Edward~L Kaplan and Paul Meier.
\newblock {Nonparametric estimation from incomplete observations}.
\newblock \emph{Journal of the American statistical association}, 53\penalty0
  (282):\penalty0 457--481, 1958.

\bibitem[Cox(1972)]{cox72}
David Cox.
\newblock {Regression Models and Life-Tables}.
\newblock \emph{Journal of the Royal Statistical Society: Series B
  (Methodological)}, 34\penalty0 (2):\penalty0 187--202, 1972.

\bibitem[Buckley and James(1979)]{BUCKLEY1979LinearData}
Jonathan Buckley and Ian James.
\newblock {Linear regression with censored data}.
\newblock \emph{Biometrika}, 66\penalty0 (3):\penalty0 429--436, 1979.

\bibitem[Katzman et~al.(2018)Katzman, Shaham, Cloninger, Bates, Jiang, and
  Kluger]{katzman2018}
Jared~L Katzman, Uri Shaham, Alexander Cloninger, Jonathan Bates, Tingting
  Jiang, and Yuval Kluger.
\newblock {DeepSurv: personalized treatment recommender system using a Cox
  proportional hazards deep neural network}.
\newblock \emph{BMC medical research methodology}, 18\penalty0 (1):\penalty0
  1--12, 2018.

\bibitem[Klein and Moeschberger(2003)]{klein2003}
John~P Klein and Melvin~L Moeschberger.
\newblock \emph{{Survival analysis: techniques for censored and truncated
  data}}, volume 1230.
\newblock Springer, 2003.

\bibitem[Breslow(1974)]{breslow1974}
Norman Breslow.
\newblock {Covariance analysis of censored survival data}.
\newblock \emph{Biometrics}, pages 89--99, 1974.

\bibitem[Wu et~al.(2018)Wu, Xiong, Yu, and Lin]{wu2018}
Zhirong Wu, Yuanjun Xiong, Stella~X Yu, and Dahua Lin.
\newblock {Unsupervised feature learning via non-parametric instance
  discrimination}.
\newblock In \emph{Proceedings of the IEEE conference on computer vision and
  pattern recognition}, pages 3733--3742, 2018.

\bibitem[He et~al.(2016)He, Zhang, Ren, and Sun]{he2016}
Kaiming He, Xiangyu Zhang, Shaoqing Ren, and Jian Sun.
\newblock {Deep residual learning for image recognition}.
\newblock In \emph{{Proceedings of the IEEE conference on computer vision and
  pattern recognition}}, pages 770--778, 2016.

\bibitem[Lederer and Martinez(2018)]{lederer2018idiopathic}
David~J Lederer and Fernando~J Martinez.
\newblock {Idiopathic pulmonary fibrosis}.
\newblock \emph{New England Journal of Medicine}, 378\penalty0 (19):\penalty0
  1811--1823, 2018.

\bibitem[Barratt et~al.(2018)Barratt, Creamer, Hayton, and
  Chaudhuri]{barratt2018idiopathic}
Shaney~L Barratt, Andrew Creamer, Conal Hayton, and Nazia Chaudhuri.
\newblock {Idiopathic pulmonary fibrosis (IPF): an overview}.
\newblock \emph{Journal of clinical medicine}, 7\penalty0 (8):\penalty0 201,
  2018.

\bibitem[Jacob et~al.(2017)Jacob, Bartholmai, Rajagopalan, Kokosi, Nair,
  Karwoski, Walsh, Wells, and Hansell]{jacob2017}
Joseph Jacob, Brian~J Bartholmai, Srinivasan Rajagopalan, Maria Kokosi, Arjun
  Nair, Ronald Karwoski, Simon~LF Walsh, Athol~U Wells, and David~M Hansell.
\newblock {Mortality prediction in idiopathic pulmonary fibrosis: evaluation of
  computer-based CT analysis with conventional severity measures}.
\newblock \emph{European Respiratory Journal}, 49\penalty0 (1), 2017.

\bibitem[Gao et~al.(2021)Gao, Kalafatis, Carlson, Pesonen, Li, Wheelock,
  Magnusson, and Sk{\"o}ld]{gao2021a}
Jing Gao, Dimitrios Kalafatis, Lisa Carlson, Ida~HA Pesonen, Chuan-Xing Li,
  {\AA}sa Wheelock, Jesper~M Magnusson, and C~Magnus Sk{\"o}ld.
\newblock {Baseline characteristics and survival of patients of idiopathic
  pulmonary fibrosis: a longitudinal analysis of the Swedish IPF Registry}.
\newblock \emph{Respiratory research}, 22\penalty0 (1):\penalty0 1--13, 2021.

\bibitem[Hofmanninger et~al.(2020)Hofmanninger, Prayer, Pan, R{\"{o}}hrich,
  Prosch, and Langs]{Hofmanninger2020}
Johannes Hofmanninger, Forian Prayer, Jeanny Pan, Sebastian R{\"{o}}hrich,
  Helmut Prosch, and Georg Langs.
\newblock {Automatic lung segmentation in routine imaging is primarily a data
  diversity problem, not a methodology problem}.
\newblock \emph{European Radiology Experimental}, 4\penalty0 (1):\penalty0
  1--13, 12 2020.

\bibitem[Nix and Weigend(1994)]{nix1994}
D.A. Nix and A.S. Weigend.
\newblock {Estimating the mean and variance of the target probability
  distribution}.
\newblock In \emph{{Proceedings of 1994 IEEE International Conference on Neural
  Networks (ICNN'94)}}, volume~1, pages 55--60 vol.1, 1994.

\bibitem[Ulyanov et~al.(2016)Ulyanov, Vedaldi, and Lempitsky]{ulyanov2016}
Dmitry Ulyanov, Andrea Vedaldi, and Victor Lempitsky.
\newblock Instance normalization: The missing ingredient for fast stylization.
\newblock \emph{arXiv preprint arXiv:1607.08022}, 2016.

\bibitem[Maas et~al.(2013)Maas, Hannun, Ng, et~al.]{maas2013}
Andrew~L Maas, Awni~Y Hannun, Andrew~Y Ng, et~al.
\newblock {Rectifier nonlinearities improve neural network acoustic models}.
\newblock In \emph{International conference on machine learning}, volume~30.
  pmlr, 2013.

\bibitem[Ioffe and Szegedy(2015)]{ioffe2015}
Sergey Ioffe and Christian Szegedy.
\newblock {Batch normalization: Accelerating deep network training by reducing
  internal covariate shift}.
\newblock In \emph{International conference on machine learning}, pages
  448--456. pmlr, 2015.

\bibitem[Loshchilov and Hutter(2018)]{loshchilov18}
Ilya Loshchilov and Frank Hutter.
\newblock {Decoupled Weight Decay Regularization}.
\newblock In \emph{{International Conference on Learning Representations}},
  2018.

\bibitem[Harrell et~al.(1996)Harrell, Lee, and Mark]{Harrell96}
Frank~E. Harrell, Kerry~L. Lee, and Daniel~B. Mark.
\newblock {Multivariable prognostic models: issues in developing models,
  evaluating assumptions and adequacy, and measuring and reducing errors}.
\newblock \emph{Statistics in Medicine}, 15\penalty0 (4):\penalty0 361--387,
  1996.

\bibitem[Agrawal et~al.(2016)Agrawal, Verma, Shah, Agarwal, and
  Sikachi]{agrawal2016}
Abhinav Agrawal, Isha Verma, Varun Shah, Abhishek Agarwal, and Rutuja~R
  Sikachi.
\newblock {Cardiac manifestations of idiopathic pulmonary fibrosis}.
\newblock \emph{Intractable \& rare diseases research}, 5\penalty0
  (2):\penalty0 70--75, 2016.

\bibitem[He et~al.(2020)He, Fan, Wu, Xie, and Girshick]{He2020a}
Kaiming He, Haoqi Fan, Yuxin Wu, Saining Xie, and Ross Girshick.
\newblock {Momentum Contrast for Unsupervised Visual Representation Learning}.
\newblock In \emph{2020 IEEE/CVF Conference on Computer Vision and Pattern
  Recognition (CVPR)}, pages 9726--9735, 2020.

\end{thebibliography}

\appendix
\section{Additional forms of censoring}
\label{app:additional_censoring}
In the main text, we focused on right-censoring, which is the most common form of censoring in survival analysis. Nevertheless, the versatility of \method enables its application to the other variants of censoring: left-censoring and interval-censoring. In this section, we discuss these additional types of censoring and delineate how our approach can be naturally adapted to handle them.
\subsection{Left-censoring}
In left-censoring, the event is known to have occurred \emph{before} a specific time $c$. Take, for example, a patient reported dead at time $c$, and this is all the information we have. We do not know the exact time of death, but we know it occurred within $\{1, \ldots, c-1\}$. This is in contrast to right-censoring, where the event is known to occur \emph{after} a particular time $c$. According to \method, we will first sample a death time $t$ from the distribution $p_\theta(t|x)$, then sample a censoring time $c$ from a distribution whose support is $\{t+1, \ldots, \tmax\}$. If we assume a uniform censoring distribution for states $c>t$, and using \eqref{eq:our_censoring_general}, a left-censored observation has the following likelihood
\beq
p(C=c|x) = \sum_{t=1}^{c-1} \frac{1}{\tmax-t} p_\theta(D=t|x)
\eeq
The likelihood for the uncensored observations remains as $p(D=t|x)$. Left-censored observations are then incorporated into the objective function as follows
\begin{align*}
\mathbb{L(\theta)} &= \sum_{i\in\uncensSet} \log p_\theta(D=t_i|x_i)\\
&\quad + \sum_{i\in\leftCensSet} \log \sum_{t=1}^{c_i-1} \frac{1}{\tmax-t} p_\theta(D=t|x) \numberthis
\end{align*}
where $\leftCensSet$ denotes the set of left-censored observations.
\subsection{Interval-censoring}
Interval-censoring arises when the event is known to have occurred \emph{within} a specific time interval $\{c_1, \ldots, c_2\}$. For instance, a patient is reported to be alive at time $c_1$ and subsequently reported dead at time $c_2$. Although the exact time of death is unknown, we know that it occurred within $\{c_1, \ldots, c_2\}$. According to our conditional censoring model, we will first sample a death time $t$ from the distribution $p_{\theta}(t|x)$, then sample a lower censoring time $c_1$ from a distribution whose support is $\{1, \ldots, t-1\}$ and an upper censoring time $c_2$ from a distribution whose support is $\{t + 1,\ldots, \tmax\}$. Similarly to the other forms of censoring, we assume a uniform censoring distribution for the states $c<t$ and $c>t$ for the two censoring distributions, respectively. The likelihood for an interval-censored observation is then
\beq
p(C_1=c_1, C_2=c_2|x) = \sum_{t=c_1+1}^{c_2-1} \frac{1}{t\br{\tmax-t}} p_\theta(D=t|x)
\eeq
The objective function is then
\begin{align*}
    \mathbb{L(\theta)} &= \sum_{i\in\uncensSet} \log p_\theta(D=t_i|x_i)\\
    &\quad + \sum_{i\in\intervalCensSet} \log \sum_{t=c_{1_i}+1}^{c_{2_i}-1} \frac{1}{t\br{\tmax-t}} p_\theta(D=t|x) \numberthis
\end{align*}
where $\intervalCensSet$ is the set of interval-censored observations.
\section{Cox with memory bank}
\label{app:coxmb}
To overcome this, we use a memory bank \cite{wu2018,He2020a} to store neural network predictions for later iterations. The memory bank, represented as $\mathcal{MB}$, is a queue of size $\lfloor K \times |\dataSet| \rfloor$ with $K$ representing the fraction of the training dataset stored, and $\lfloor . \rfloor$ representing the floor function. A $K$ value of 1 corresponds to the storage of the entire training set in the memory bank, while a $K=0$ means that no samples are stored, which is equivalent to the standard Cox objective. For every training iteration $i$, we calculate predictions $g_{\theta^i}(x^i)$ for the $i$-th minibatch $m^i$ and store them in $\mathcal{MB}$, along with the corresponding event indicators $\delta^i$ and death times $t^i$ (or censoring times $c^i$ for censored samples). The memory bank $\mathcal{MB}$ is updated as
\beq
\mathcal{MB} \leftarrow \mathcal{MB} \mathbin\Vert \cb{g_{\theta^i}(x^i), \delta^i, t^i, c^i}
\eeq
where $\mathbin\Vert$ denotes concatenation. If the memory bank is full (\ie $|\mathcal{MB}| = \lfloor K \times |\dataSet| \rfloor$), the oldest samples are removed and new samples are added. After $I$ iterations, $\mathcal{MB}$ will contain the tuples $\{g_{\theta^i}(x^i), \delta^i, t^i, c^i\}_{i=1}^I$. At each iteration $i$, we calculate the risk set $R_n^i$ for each uncensored patient $n$ in $\mathcal{MB}$ using the stored event indicators and times. The Cox loss for samples in $\mathcal{MB}$ is then clalculated using the risk set $R_n^i$ and the available predictions in the memory bank as
\beq
L(\theta^i) \equiv \frac{1}{\mathcal{N}_{\text{uncensMB}}^i} \sum_{n \in {\mathcal{N}_{\text{uncensMB}}^i}}  \sq{g_{\mathcal{MB}_n}(\theta^{\leq i}) - \log{\sum_{m \in R_n^i}\exp(g_{\mathcal{MB}_m}(\theta^{\leq i}))}}
\eeq
where $\mathcal{N}_{\text{uncensMB}}^i$ is the set of uncensored samples in $\mathcal{MB}$ at iteration $i$, and $g_{\mathcal{MB}_n}(\theta^{\leq i})$ and $g_{\mathcal{MB}_m}(\theta^{\leq i})$ are the predictions for patients $n$ and $m$ in $\mathcal{MB}$, respectively, and are functions of the model parameters at iteration $i$ or any previous iteration $<i$. The loss is used to update the current parameters of the model $\theta^i$. By updating $\mathcal{MB}$ at each iteration and using it to calculate the loss, we can effectively approximate the Cox loss on a sample size larger than allowed by the standard Cox objective. We refer to this method as CoxMB and compare its performance to the standard Cox objective in our experiments.
\color{black}

\newpage
\section{Model architecture}
\label{app:model_architecture}
\begin{figure}[!h]
    \centering
    \includegraphics[width=0.549\linewidth]{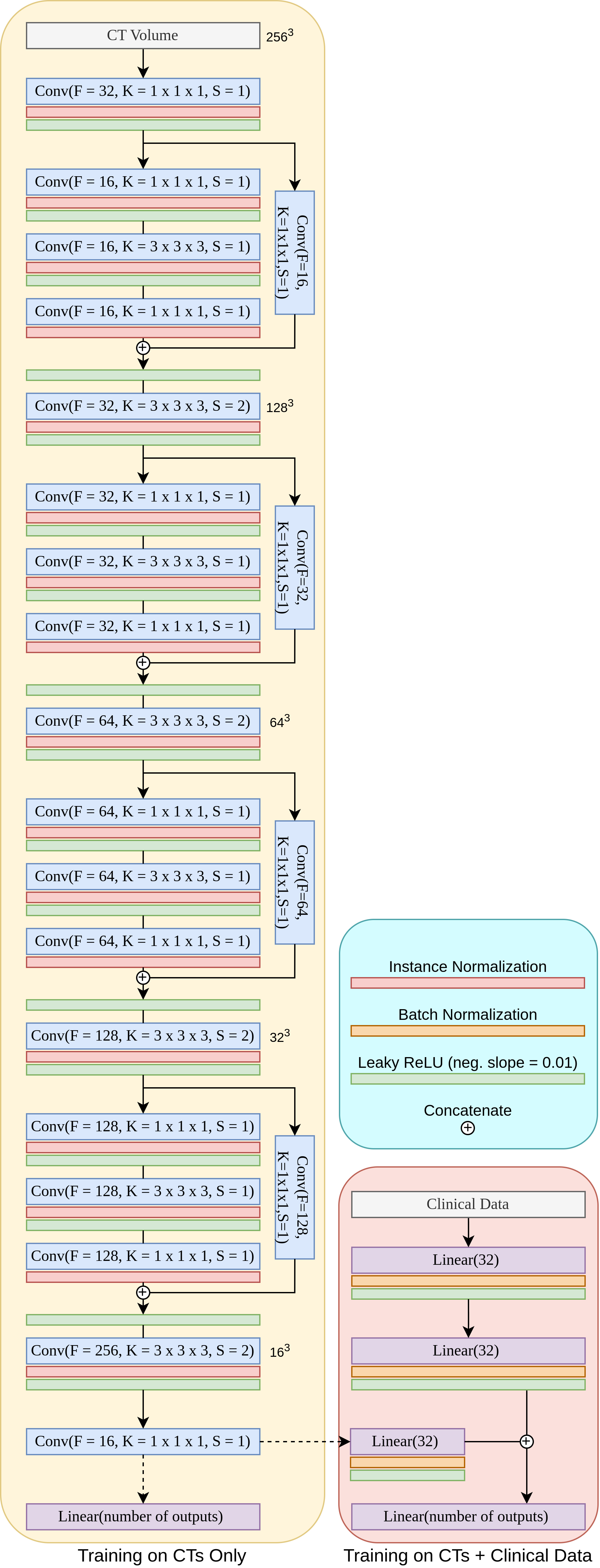}
    \caption{Model architecture. Left: 3D CNN for processing HRCT scans. Right: MLP to process clinical data. $F$: Number of filters, $K$: kernel size, $S$: stride.
    In the case of using HRCTs only, the architecture on the left is used. In the case of using HRCT and clinical data, the outputs of CNN and MLP are concatenated.}
\end{figure}

\section{Effect of memory bank size in CoxMB}
We examine the effect of the size of the memory bank in the CoxMB model, trained on imaging data. $K$ is the fraction of training samples stored in the memory bank during training. We train the CoxMB model with different values of $K$ and report the results in \tabref{tab:memory_bank_size}. We observe that the performance of the CoxMB model improves as the memory bank size increases. This is expected, as a larger memory bank allows the model to store more information about the ranking of patients' survival times, which is then used to penalise the model for inaccuracies in predicting the ranking. We anticipate that this depends on the size of the training set and thus requires tuning for each dataset. However, we observe that the performance of the CoxMB model is relatively stable for a wide range of $K$ values (0.4 to 1.0), suggesting that the model is not very sensitive to the choice of $K$.

\begin{table}[h]
    \centering
    \caption{Effect of memory bank size on the performance of CoxMB model.}
    \label{tab:memory_bank_size}
    \begin{tabular}{@{}cc@{}}
        \toprule
        K & C-Index \\ 
        \midrule
        0.0 & 67.441 $\pm$ 4.572 \\ 
        0.2 & 67.968 $\pm$ 2.712 \\ 
        0.4 & 70.884 $\pm$ 3.844 \\ 
        0.6 & 70.154 $\pm$ 0.975 \\ 
        0.8 & \textbf{73.294} $\pm$ \textbf{4.056} \\ 
        1.0 & 71.067 $\pm$ 5.572 \\ 
        \bottomrule
    \end{tabular}
\end{table}

\end{document}